\documentclass{article} 
\usepackage{iclr2027_conference,times}

\usepackage{amsmath,amsfonts,bm}

\def\eqref#1{equation~\ref{#1}}

\def\1{\bm{1}}

\DeclareMathAlphabet{\mathsfit}{\encodingdefault}{\sfdefault}{m}{sl}
\SetMathAlphabet{\mathsfit}{bold}{\encodingdefault}{\sfdefault}{bx}{n}

\usepackage{hyperref}
\usepackage{url}
\usepackage{graphicx}
\usepackage{booktabs}
\usepackage{multirow}
\usepackage[table,dvipsnames]{xcolor}
\definecolor{lightgreen}{HTML}{00C707}
\definecolor{tarsred}{HTML}{C00000}
\definecolor{darkgreen}{HTML}{008000}
\definecolor{lightgray}{gray}{0.9}
\definecolor{headgray}{gray}{0.92}

\usepackage{float}
\usepackage{placeins}

\title{TARS: Extending Reinforcement Learning Gains to the Long Tail under Limited Data for Domain-Specific Visual Recognition}

\author{Beomchan Park\thanks{Equal contribution.}\hspace{0.4em},\hspace{0.3em}
Seongho Kim\footnotemark[1]\hspace{0.4em} \& Yong Man Ro\thanks{Corresponding author.} \\
School of Electrical Engineering\\
KAIST\\
Yuseong-gu, Daejeon, Republic of Korea \\
\texttt{\{bpark0810, taho43, ymro\}@kaist.ac.kr}
}

\iclrfinalcopy 
\begin{document}

\maketitle
\begin{abstract}
Visual recognition in domain-specific applications, such as biodiversity research and agriculture, often suffers from limited and imbalanced training data. Large Vision-Language Models (LVLMs) offer a promising starting point with rich visual and semantic knowledge acquired through web-scale pretraining, but naive supervised fine-tuning (SFT) on such data can cause them to overfit to head classes. Although reinforcement learning (RL) is reported to generalize better than SFT, our analysis shows that its gains under Group Relative Policy Optimization (GRPO) do not consistently extend to tail classes. To address this limitation, we propose \textbf{Tailness-aware Advantage ReScaling (TARS)}, a novel RL-based method that amplifies GRPO advantages for classes that are under-represented and poorly recognized. This amplification is guided by \textit{tailness}, defined as the product of class rarity in the training data and the per-class error rate of the SFT model. Experiments on four domain-specific classification and detection benchmarks show that TARS consistently improves tail-class performance over SFT and vanilla GRPO, while achieving competitive results on tail classes and the highest overall performance among the compared GRPO methods. TARS also provides further gains when applied on top of imbalance-aware SFT methods, demonstrating that the two approaches are complementary.
\end{abstract}

\section{Introduction}

Visual recognition in domain-specific applications, such as classifying species for biodiversity research \citep{wah2011caltech, van2018inaturalist} and identifying pests or plant diseases in agriculture \citep{wu2019ip102, singh2020plantdoc}, often suffers from limited data and class imbalance. Annotation requires domain expertise, which makes collecting large annotated datasets difficult. Moreover, some classes are encountered far more frequently than others during data collection in real-world settings, resulting in imbalanced training sets. Under these constraints, the objective is to fully leverage the scarce data while achieving balanced performance across classes along with high overall performance. 
\looseness=-1

Recent Large Vision-Language Models (LVLMs) have shown strong capabilities after task-specific fine-tuning \citep{liu2024improved, bai2025qwen2}. Research on these models extends beyond multimodal reasoning and comprehension \citep{yue2024mmmu, liu2024mmbench} to visual recognition tasks such as classification and detection \citep{zhang2024visually, yin2025rod, conti2025large, li2025lmm, zang2025contextual, Kisel_2026_CVPR}. In this setting, LVLMs perform visual recognition by autoregressively generating class names and object coordinates, allowing recognition tasks to leverage their pretrained visual and semantic knowledge. Although LVLMs may struggle with domain-specific concepts out of the box, their pretrained knowledge makes them promising starting points for domain-specific visual recognition under limited data.

\begin{figure}[t!]
\begin{center}
\includegraphics[width=\linewidth]{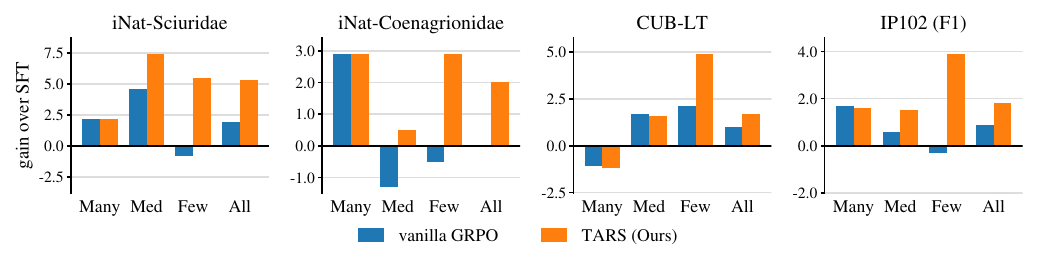}
\vspace{-9.0mm}
\end{center}
\caption{Gain over the CE loss SFT baseline, broken down by class frequency in the training data. Vanilla GRPO does not consistently improve performance on tail classes and sometimes underperforms SFT, while TARS achieves higher tail-class and overall performance than both SFT and vanilla GRPO on every benchmark. IP102 is reported in F1@0.5 and the others in accuracy.}
\label{fig:fig1}
\vspace{-4.0mm}
\end{figure}

However, supervised fine-tuning (SFT) learns by imitating target responses and can overfit to head classes under limited and imbalanced training data. Even with existing imbalance-aware losses and sampling methods, substantial performance gaps between head and tail classes remain (Tabs.~\ref{tab:open_world_classification}--\ref{tab:ip102_detection}). Reinforcement learning (RL) allows models to explore responses guided by verifiable rewards and has been reported to generalize better than SFT \citep{chu2025sft}, making it a promising approach to further improving generalization after SFT. In particular, Group Relative Policy Optimization (GRPO) \citep{shao2024deepseekmath} has demonstrated improvements in LVLM-based visual recognition using correctness and localization rewards \citep{liu2025visual}. These findings motivate investigating GRPO as a complementary optimization stage to improve recognition of tail classes from limited training examples. Yet, as shown in Fig.~\ref{fig:fig1}, vanilla GRPO does not consistently improve tail-class performance and even falls below the SFT baseline in several cases.

To understand this limitation, it is useful to consider two properties of GRPO. First, although the policy explores alternative responses, the objective of GRPO is still an expectation over queries drawn from the training data, so frequent classes contribute to policy updates more often than rare classes. Second, GRPO computes advantages by normalizing rewards across rollouts of the same query. This within-query normalization does not explicitly account for class frequency or aggregate class-level recognition performance. This motivates identifying classes with both limited exposure and substantial recognition error and amplifying the learning signals they produce.
\looseness=-1

To this end, we propose \textbf{Tailness-aware Advantage ReScaling (TARS)}, a novel RL-based generalization method that rescales GRPO advantages based on per-class \textit{tailness} for domain-specific visual recognition. Tailness is a measure designed to identify classes that need stronger learning signals. Class frequency is a natural starting point, but frequency alone may not reflect how well the model recognizes each class \citep{sinha2023difficulty, Wang_2021_CVPR}. For example, a tail class with distinctive visual features may already be well recognized. We therefore define tailness as the product of class rarity in the training data and the per-class error rate of the SFT model, accounting for both under-representation and recognition difficulty. TARS then selectively amplifies GRPO advantages based on class tailness. The same formulation applies across visual recognition tasks, including classification and detection.
\looseness=-1

We evaluate TARS on four domain-specific benchmarks with limited and imbalanced training data: iNat-Sciuridae and iNat-Coenagrionidae \citep{van2018inaturalist} and CUB-LT \citep{samuel2021generalized} for classification, and IP102 \citep{wu2019ip102} for detection. TARS consistently improves tail-class performance over SFT and vanilla GRPO, and achieves strong tail-class and overall performance compared with competitive GRPO baselines \citep{kumar2026dive, zhou2025disco, dai2026qoq}. Also, applying TARS on top of imbalance-aware SFT methods \citep{lin2017focal, du2023global, kang2020decoupling} yields further gains, demonstrating their complementarity.
\looseness=-1

Our contributions are summarized as follows:
\begin{itemize}
    \item We show that the generalization gains of vanilla GRPO do not consistently extend to tail classes in domain-specific visual recognition under limited and imbalanced training data.
    \item We propose \textbf{Tailness-aware Advantage ReScaling (TARS)}, a novel RL-based generalization method that selectively amplifies GRPO advantages using tailness, defined as the product of class rarity and the per-class error rate of the SFT model.
    \item Across four domain-specific classification and detection benchmarks, TARS improves tail-class performance over SFT and vanilla GRPO and achieves strong tail-class and overall performance compared with competitive GRPO baselines. Further gains on top of imbalance-aware SFT methods demonstrate their complementarity with TARS.
\end{itemize}

\section{Related Works}

\subsection{Visual Recognition with LVLMs}
Recent studies evaluate the visual recognition capabilities of LVLMs and identify limitations in their out-of-the-box performance \citep{zhang2024visually, Kisel_2026_CVPR}. Challenges include generating category names without a candidate list in open-world recognition \citep{conti2025large} and connecting visual evidence to the correct concept name \citep{he2025finedefics}. To improve recognition, a line of work adapts LVLMs to classification and detection through supervised fine-tuning on task-formatted data \citep{yin2025rod, li2025lmm, zang2025contextual, zhan2025griffon}. Moreover, Visual-RFT \citep{liu2025visual} applies GRPO to visual recognition using verifiable accuracy- and IoU-based rewards, demonstrating gains in few-shot classification and detection. Subsequent studies \citep{yu2026perception, liu2026visionreasoner, zhan2026visionr1, jiang2026detect} further investigate RL-based post-training for various visual perception objectives. Our work focuses on extending GRPO's generalization gains to tail classes in domain-specific visual recognition under limited and imbalanced training data. Existing supervised approaches to class imbalance include loss reweighting \citep{lin2017focal, du2023global, sinha2022class, sinha2023difficulty} or class-balanced sampling \citep{kang2020decoupling}. However, substantial performance gaps between head and tail classes remain even with these methods, motivating further improvement through GRPO. Unlike supervised reweighting approaches that adjust learning emphasis during SFT, TARS operates during GRPO-based LVLM post-training, selectively amplifying the advantages of under-represented classes primarily based on class rarity, with recognition error providing additional information to refine this emphasis.
\looseness=-1

\subsection{Reinforcement Learning for Generalization}
RL with verifiable rewards has shown generalization benefits over SFT \citep{chu2025sft}, motivating research on how to learn more effectively from problems that remain challenging for the model. In mathematical reasoning, GRPO-LEAD \citep{zhang2025grpo} and DGPO \citep{dai2026harder} prioritize learning from questions with lower rollout success rates. Similarly, UniVG-R1 \citep{bai2025univg} assesses the difficulty of individual visual grounding queries from their average grounding rewards across rollouts and prioritizes harder queries. Another line of work seeks to improve performance on under-represented domains during joint training. DISCO \citep{zhou2025disco} combines domain-frequency weighting with prompt-level difficulty across tasks such as mathematical reasoning, question answering, and sentiment classification. DRPO \citep{dai2026qoq} balances learning across clinical domains in multimodal medical reasoning based on domain frequency and reward statistics within each training batch. TARS seeks to extend RL's generalization gains to tail classes in domain-specific visual recognition under limited and imbalanced training data. It combines class rarity in the full training set with the SFT model's recognition error aggregated over the examples of each class. The resulting tailness determines how strongly GRPO advantages are amplified for each class.
\looseness=-1

\section{Method}
\begin{figure}[t!]
    \begin{center}
    \includegraphics[width=\linewidth]{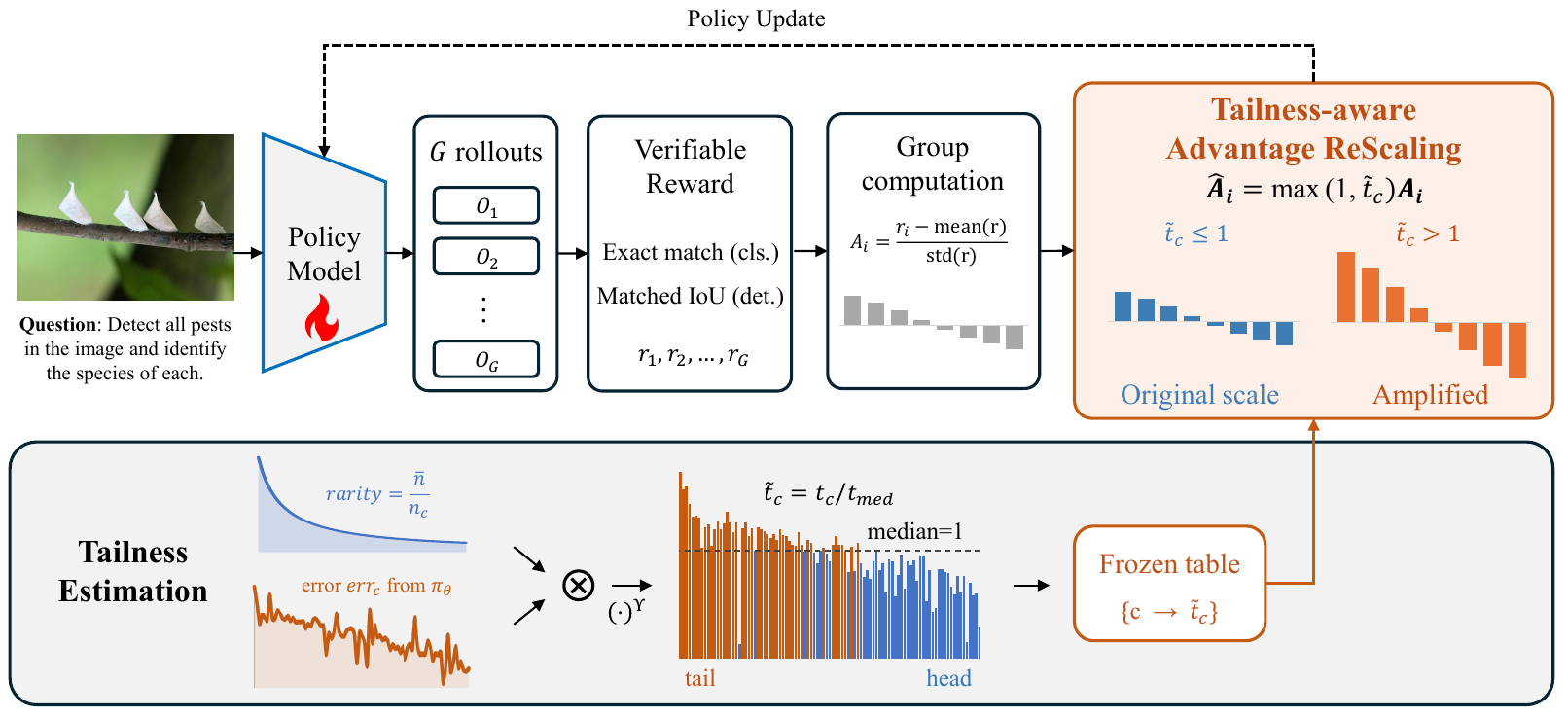}
    \end{center}
    \vspace{-4.0mm}
    \caption{Overview of TARS. Class tailness is computed once before RL from class rarity and the SFT model's recognition error, then normalized by its median across classes. During RL, these fixed scores rescale GRPO advantages through $\max(1,\tilde{t}_c)$, amplifying signals for classes with above-median tailness while retaining the original scale for the rest.}
    \label{fig:overall}
\end{figure}

We consider domain-specific visual recognition over a set of classes $\mathcal{C}$, given a training set $\mathcal{D}$ that is limited in size and imbalanced over $\mathcal{C}$. Our RL stage starts from an SFT checkpoint and further optimizes the LVLM with GRPO, aiming for accurate recognition across all classes rather than only the frequent ones. We first review GRPO and discuss its limitations under class imbalance. We then introduce tailness and TARS, followed by the task-dependent verifiable rewards used in our experiments. Fig.~\ref{fig:overall} summarizes the resulting procedure.

\subsection{Preliminaries}
Group Relative Policy Optimization (GRPO) \citep{shao2024deepseekmath} optimizes a policy without a learned value function by estimating advantages from the rewards of multiple responses to the same query. For a query $q$ drawn from $\mathcal{D}$, the policy $\pi_{\theta_{\mathrm{old}}}$ samples a group of $G$ responses $\{o_i\}_{i=1}^{G}$, each of which receives a scalar reward $r_i$ from a verifiable reward rule. The advantage of a response is obtained by standardizing the rewards within the group,
\begin{equation}
A_i =
\frac{r_i - \mathrm{mean}(\{r_j\}_{j=1}^{G})}
{\mathrm{std}(\{r_j\}_{j=1}^{G})+\epsilon},
\label{eq:adv}
\end{equation}
where $\epsilon>0$ is a tiny positive constant for numerical stability.

For clarity, we present a simplified response-level formulation, omitting token-level details. For each response, GRPO forms the clipped surrogate \citep{schulman2017proximal}
\begin{equation}
\ell_i(\theta) = \min \big(\rho_i A_i,\, \mathrm{clip}(\rho_i,\,1-\varepsilon,\,1+\varepsilon)A_i \big),
\label{eq:surrogate}
\end{equation}
where $\rho_i = \pi_\theta(o_i \mid q) / \pi_{\theta_{\mathrm{old}}}(o_i \mid q)$ is the importance ratio and $\varepsilon$ sets the clipping range. Maximizing this surrogate encourages responses with positive advantages and discourages those with negative advantages. GRPO then maximizes
\begin{equation}
\mathcal{J}_{\mathrm{GRPO}}(\theta) = 
\mathbb{E}_{q,\,\{o_i\}} \left[
\frac{1}{G}\sum_{i=1}^{G}\ell_i(\theta)
-\beta D_{\mathrm{KL}} \big[\pi_\theta \,\|\, \pi_{\mathrm{ref}}\big]
\right],
\label{eq:grpo}
\end{equation}
where the expectation is over $q\sim\mathcal{D}$ and $\{o_i\}_{i=1}^{G}\sim\pi_{\theta_{\mathrm{old}}}(\cdot\mid q)$, and $\beta$ weights the divergence from the reference policy $\pi_{\mathrm{ref}}$.

Eq.~\ref{eq:adv} highlights a limitation central to this work. Although GRPO's relative advantages implicitly reflect the model's recognition capability at the \textit{query level}, they do not explicitly account for \textit{class} frequency or aggregate \textit{class-level} recognition performance. Since queries are uniformly sampled from $\mathcal{D}$, frequent classes contribute to the objective more often, while poorly recognized tail classes receive no targeted amplification to compensate for their limited exposure. This can leave the LVLM's recognition capabilities biased toward head classes, motivating class-aware advantage weighting to counteract the imbalance in class exposure.
\looseness=-1

\subsection{Tailness-aware Advantage ReScaling}
\paragraph{Measuring tailness.} We define tailness to identify classes that need stronger learning signals under imbalanced training data. Class rarity captures limited exposure, but a rare class may already be well recognized. We therefore combine rarity with the SFT model's per-class recognition error to account for how well each class is already learned.

Let $n_c$ be the number of training queries in which class $c$ occurs and let $\bar{n} = \frac{1}{|\mathcal{C}|}\sum_{c \in \mathcal{C}} n_c$ be the average occurrence count. Both classification and detection are posed as one query per image (see Tab.~\ref{tab:query-examples} for examples). Thus, $n_c$ counts images labeled as $c$ for classification and images containing at least one instance of $c$ for detection. This measures class exposure at the query level rather than the instance level.

\begin{table}[t]
\setlength{\tabcolsep}{4pt}
\caption{Query examples for classification and detection.}
\label{tab:query-examples}
\begin{center}
\begin{tabular}{@{}p{0.15\columnwidth}p{0.75\columnwidth}@{}}
\toprule
Task & Query prompt \\
\midrule
Classification
& What is the species of the \textit{squirrel / damselfly / bird} shown in this image? \\
\midrule
Detection
& Detect all \textit{pests} in the image and identify the species of each. \\
\bottomrule
\end{tabular}
\end{center}
\vspace{-3.0mm}
\end{table}

Before RL begins, we run a single inference pass over the training data with the initial SFT policy $\pi_{\theta_0}$ and measure its per-class performance $s_c \in [0,1]$ using the same metric as at evaluation time. In our experiments, we use accuracy for classification and F1 score at an IoU threshold of $0.5$ for detection. We define the recognition error as $err_c = 1 - s_c$ and the raw tailness of class $c$ as
\begin{equation}
t_c = \left( \frac{\bar{n}}{n_c} \cdot err_c \right)^{\gamma},
\label{eq:tailness}
\end{equation}
where $\bar{n}/n_c$ is the rarity of $c$ relative to the average occurrence count and $\gamma > 0$ controls the contrast across classes. The product assigns greater tailness to rarer classes at the same error level, while reducing it for better-recognized classes at the same frequency. Setting $\gamma<1$ compresses the relative differences between positive values, whereas $\gamma>1$ accentuates them, in both cases preserving the ordering across classes.

\paragraph{Calibration.} The scale of raw tailness varies across datasets, so we normalize it by the median tailness across classes,
\begin{equation}
\tilde{t}_c = \frac{t_c}{\max(t_{med},\epsilon)},
\qquad
t_{med} = \mathrm{median}\big(\{t_c\}_{c \in \mathcal{C}}\big),
\label{eq:calib}
\end{equation}
where $\epsilon > 0$ is a tiny positive constant to prevent division by zero. Thus, $\tilde{t}_c>1$ indicates above-median tailness, reflecting both class rarity and recognition error. We use the median because it is less sensitive than the mean to extremely large raw tailness values.

\paragraph{Rescaling the advantage.} For a query $q$ associated with a single class $c(q)$, TARS multiplies its GRPO advantages by the class weight $w_c=\max(1,\tilde{t}_c)$ with $c=c(q)$. Since this weight is positive, scaling the advantages is equivalent to scaling the clipped surrogate in Eq.~\ref{eq:surrogate}, yielding
\begin{equation}
\begin{aligned}
\mathcal{J}_{\mathrm{TARS}}(\theta) =
\mathbb{E}_{q,\,\{o_i\}}
\Big[\frac{1}{G} \sum_{i=1}^{G}
\textcolor{tarsred}{\max\big(1,\ \tilde{t}_{c(q)}\big)}\, \ell_i(\theta)
- \beta\, D_{\mathrm{KL}}\big[\pi_\theta \,\|\, \pi_{\mathrm{ref}}\big] \Big].
\end{aligned}
\label{eq:tars}
\end{equation}
The scaling is one-sided. Classes with $\tilde{t}_c \le 1$ retain the original GRPO advantage scale, while those with $\tilde{t}_c > 1$ receive amplification proportional to their calibrated tailness. This preserves the advantage signs and clipping conditions.


Rare classes contribute to policy updates less frequently because they are sampled less often from $\mathcal{D}$. To compensate for this imbalance, TARS uses the inverse-frequency factor in Eq.~\ref{eq:tailness} to amplify their gradient contributions when they appear. For amplified single-class queries, $\gamma=1$ fully offsets the sampling-frequency factor, while $0<\gamma<1$ partially offsets it. Recognition error further adjusts the emphasis according to how well each class is already recognized by the SFT model.

Since tailness remains fixed throughout RL, $\nabla_\theta[w_c\ell_i(\theta)]=w_c\nabla_\theta\ell_i(\theta)$. Each response's surrogate gradient therefore retains its direction, while reweighting the contributions of different classes can change the overall gradient direction. The KL term is not rescaled, reducing its relative influence for queries with larger weights.

\paragraph{Queries with multiple classes.} In classification, each query corresponds to a single class. In detection, an image may contain multiple classes. Let $\mathcal{C}(q)$ denote the set of distinct classes present in the ground-truth annotations for query $q$. We average their weights,
\begin{equation}
w(q) = \frac{1}{|\mathcal{C}(q)|} \sum_{c \in \mathcal{C}(q)}
       \max\!\big(1, \tilde{t}_c\big),
\label{eq:agg}
\end{equation}
and use $w(q)$ in place of $\max(1,\tilde{t}_{c(q)})$ in Eq.~\ref{eq:tars}. This assigns a single amplification factor to the query based on the tailness of the multiple classes present in the image. The resulting query weight remains positive and fixed, preserving the scaling properties described above.

\subsection{Verifiable Rewards for Visual Recognition}
Following the use of verifiable rewards in Visual-RFT \citep{liu2025visual}, we compute rollout rewards from classification correctness and, for detection, class-label agreement and bounding-box overlap. Prior studies show that explicit reasoning does not consistently improve visual perception performance and that training without reasoning traces can achieve comparable or better results \citep{yu2026perception, li2026think}. Accordingly, we adopt direct-answer generation without explicit reasoning traces and do not use a separate think-format reward. For both tasks, labels are converted to lowercase with separators removed before exact matching.


\paragraph{Classification.}
The model is asked to predict the class label of the image. The reward is an exact match between the normalized prediction $\hat{y}$ and ground-truth label $y$,
\begin{equation}
r = \textbf{1}\big[\, \hat{y} = y \,\big].
\label{eq:reward-cls}
\end{equation}
Any response that cannot be parsed into a class label receives zero.

\paragraph{Detection.}
The prompt asks the model to locate objects in the image and predict their class labels. The response is a JSON list in which each item contains a bounding box and a class label.

Let $\hat{\mathcal{B}} = \{(\hat{b}_k, \hat{y}_k)\}$ be the predicted set and $\mathcal{B} = \{(b_j, y_j)\}$ the annotated set. A predicted item is eligible to match an annotated item only if their normalized labels agree and their IoU is positive. Eligible pairs are matched greedily in descending order of IoU, with each item used at most once, yielding a matching $\mathcal{M}$. The reward is the total matched overlap divided by the larger of the two set sizes, \looseness=-1
\begin{equation}
r = \frac{1}{\max\big(|\hat{\mathcal{B}}|,\, |\mathcal{B}|\big)}
    \sum_{(k,j) \in \mathcal{M}} \mathrm{IoU}\big(\hat{b}_k,\, b_j\big).
\label{eq:reward-det}
\end{equation}
Matched boxes receive credit proportional to their IoU, while unmatched predictions and annotations remain included in the counts used for normalization. Unlike evaluation with F1 score at the IoU threshold of $0.5$, the reward gives partial credit to matched boxes even below this threshold, reflecting differences in localization quality. A response that cannot be parsed into the required JSON form receives zero. We use no separate output-format reward since the SFT checkpoints already adhere well to the required JSON format.

\section{Experiments}
\label{sec:exp}

\subsection{Experimental Setup}
\paragraph{Datasets.}
We evaluate TARS on four benchmarks covering diverse visual characteristics across domain-specific applications in biodiversity and agriculture, spanning both classification and detection. To study visual recognition under limited and imbalanced training data, we select benchmarks with at most 20,000 training images and an imbalance factor (IF) greater than 10, defined as the ratio of the largest to the smallest class frequency in the training set. For classification, we use two family-level subsets of iNaturalist-2018 \citep{van2018inaturalist}, iNat-Sciuridae and iNat-Coenagrionidae, and CUB-LT \citep{samuel2021generalized}, derived from CUB-200-2011 \citep{wah2011caltech}, all from the biodiversity domain. For detection, we use IP102 \citep{wu2019ip102} for agricultural pest recognition. Dataset statistics are provided in Appendix~\ref{app:dataset_details}.


\paragraph{Evaluation Protocol.}
All benchmarks use an open-world setting in which the model generates class names as free-form text without access to a candidate label list \citep{zhang2024visually}. Following \citet{liu2019large}, we divide classes by their training image count $n_c$ into Many ($n_c > 100$), Medium ($20 \leq n_c \leq 100$), and Few ($n_c < 20$) for IP102 and the iNaturalist subsets. For CUB-LT, whose largest class has fewer than 100 training images, we use Many ($n_c > 20$), Medium ($5 \leq n_c \leq 20$), and Few ($n_c < 5$). For classification, we report accuracy based on exact matching between normalized predictions and ground-truth labels. For detection, the model predicts boxes and class labels without confidence scores, so we report precision, recall, and F1 score following \citet{jiang2026detect}. Within each image, predictions and ground-truth boxes of the same class are matched one-to-one using Hungarian matching at IoU $\geq 0.5$. 
\looseness=-1

\paragraph{Implementation Details.}
We adopt Qwen2.5-VL-3B-Instruct \citep{bai2025qwen2} as our base LVLM and update all parameters except the vision encoder in both SFT and RL. Each GRPO optimization step uses 8 queries with group size $G=8$, yielding a batch of 64 rollouts. We set the KL coefficient to $\beta=0.04$ and the sampling temperature to 1.0. The tailness exponent is $\gamma=0.75$ for classification and $\gamma=1.0$ for detection. We repeat each RL run with three random seeds and report the mean. All experiments use bfloat16 precision on 4 NVIDIA H100 GPUs. Further implementation details are provided in Appendix~\ref{app:additional}.

\subsection{Main Results}
We compare TARS with vanilla GRPO on top of four SFT methods, including standard cross-entropy loss, focal loss \citep{lin2017focal}, class-frequency reweighting \citep{du2023global}, and class-balanced sampling \citep{kang2020decoupling}. For each SFT method, TARS and vanilla GRPO start from the same checkpoint and use the same RL settings. Focal loss shifts learning emphasis toward harder predictions, while class-frequency reweighting and class-balanced sampling increase the contribution of under-represented classes through loss weighting and data sampling, respectively. These comparisons test whether TARS provides additional benefits beyond established approaches that address class imbalance during supervised recognition. 

\begin{table}[t!]
    \setlength{\tabcolsep}{1.5pt}
    \newcolumntype{M}{>{\centering\arraybackslash}m{3.2em}}
    \caption{
    Domain-specific classification accuracy (\%) on iNat-Sciuridae, iNat-Coenagrionidae, and CUB-LT. Many, Medium, and Few group classes by training frequency, and All denotes overall accuracy. The first row shows Qwen2.5-VL-3B-Instruct without task-specific fine-tuning. SFT baselines use cross-entropy (CE Loss), focal loss (Focal Loss), class-frequency reweighting (Reweighting), or class-balanced sampling (Sampling). Bold and underlined values mark the best and second-best results in each column within each SFT block. \textcolor{tarsred}{Red} subscripts show percentage-point gains over the corresponding SFT baseline for Few and All.
    }
    \vspace{-2.0mm}
    \label{tab:open_world_classification}
    \begin{center}
    \resizebox{\textwidth}{!}{%
    \begin{tabular}{l *{12}{M}}
        \toprule
        \multirow{2}{*}{\textbf{Method}}
        & \multicolumn{4}{c}{\textbf{iNat-Sciuridae}}
        & \multicolumn{4}{c}{\textbf{iNat-Coenagrionidae}}
        & \multicolumn{4}{c}{\textbf{CUB-LT}} \\
        \cmidrule(lr){2-5}
        \cmidrule(lr){6-9}
        \cmidrule(lr){10-13}
        & Many & Med & Few & All
        & Many & Med & Few & All
        & Many & Med & Few & All \\
        \midrule
        Qwen2.5-VL-3B
        & 30.0 & 11.1 & 7.1 & 14.8
        & 24.4 & 18.1 & 8.7 & 16.1
        & 4.9 & 5.9 & 3.7 & 5.2 \\
        \midrule
        CE Loss
        & \underline{86.7} & 50.0 & \underline{42.9} & 57.4
        & \underline{77.8} & \underline{56.9} & \underline{43.5} & \underline{57.0}
        & \textbf{92.7} & 77.9 & 65.8 & 78.2 \\
        + Vanilla
        & \textbf{88.9} & \underline{54.6} & 42.1 & \underline{59.3}
        & \textbf{80.7} & 55.6 & 43.0 & \underline{57.0}
        & \underline{91.6} & \textbf{79.6} & \underline{67.9} & \underline{79.2} \\
        \rowcolor{lightgray}
        \textbf{+ TARS (Ours)}
        & \textbf{88.9} & \textbf{57.4} & \textbf{48.4}$_{\textcolor{tarsred}{\uparrow 5.5}}$ & \textbf{62.7}$_{\textcolor{tarsred}{\uparrow 5.3}}$
        & \textbf{80.7} & \textbf{57.4} & \textbf{46.4}$_{\textcolor{tarsred}{\uparrow 2.9}}$ & \textbf{59.0}$_{\textcolor{tarsred}{\uparrow 2.0}}$
        & 91.5 & \underline{79.5} & \textbf{70.7}$_{\textcolor{tarsred}{\uparrow 4.9}}$ & \textbf{79.9}$_{\textcolor{tarsred}{\uparrow 1.7}}$ \\
        \midrule
        Focal Loss
        & 86.7 & \textbf{58.3} & \underline{45.2} & \underline{61.1}
        & 75.6 & \underline{56.9} & \underline{44.9} & \underline{57.0}
        & 91.8 & 78.2 & \underline{68.3} & \underline{78.7} \\
        + Vanilla
        & \textbf{90.0} & 53.7 & 40.5 & 58.6
        & \textbf{77.8} & 53.5 & 42.8 & 55.4
        & \textbf{92.3} & \underline{78.3} & 67.4 & \underline{78.7} \\
        \rowcolor{lightgray}
        \textbf{+ TARS (Ours)}
        & \underline{88.9} & \underline{57.4} & \textbf{47.6}$_{\textcolor{tarsred}{\uparrow 2.4}}$ & \textbf{62.3}$_{\textcolor{tarsred}{\uparrow 1.2}}$
        & \underline{76.7} & \textbf{57.0} & \textbf{47.9}$_{\textcolor{tarsred}{\uparrow 3.0}}$ & \textbf{58.3}$_{\textcolor{tarsred}{\uparrow 1.3}}$
        & \underline{92.0} & \textbf{78.7} & \textbf{69.9}$_{\textcolor{tarsred}{\uparrow 1.6}}$ & \textbf{79.4}$_{\textcolor{tarsred}{\uparrow 0.7}}$ \\
        \midrule
        Reweighting
        & 86.7 & \underline{58.3} & \underline{45.2} & 61.1
        & \underline{77.8} & \textbf{61.1} & \underline{44.9} & \underline{59.1}
        & 91.7 & 78.6 & \underline{69.2} & \underline{79.1} \\
        + Vanilla
        & \underline{90.0} & \underline{58.3} & 43.6 & \underline{61.4}
        & \textbf{82.2} & 56.9 & 43.0 & 57.9
        & \textbf{92.6} & \underline{79.2} & 67.0 & 79.0 \\
        \rowcolor{lightgray}
        \textbf{+ TARS (Ours)}
        & \textbf{91.1} & \textbf{62.9} & \textbf{49.2}$_{\textcolor{tarsred}{\uparrow 4.0}}$ & \textbf{65.4}$_{\textcolor{tarsred}{\uparrow 4.3}}$
        & \textbf{82.2} & \underline{57.9} & \textbf{48.3}$_{\textcolor{tarsred}{\uparrow 3.4}}$ & \textbf{60.2}$_{\textcolor{tarsred}{\uparrow 1.1}}$
        & \underline{92.1} & \textbf{79.7} & \textbf{72.0}$_{\textcolor{tarsred}{\uparrow 2.8}}$ & \textbf{80.4}$_{\textcolor{tarsred}{\uparrow 1.3}}$ \\
        \midrule
        Sampling
        & \textbf{90.0} & 58.3 & \underline{54.8} & 65.7
        & 73.3 & 52.8 & 43.5 & 54.3
        & 89.6 & \textbf{80.8} & \underline{73.9} & \underline{81.0} \\
        + Vanilla
        & \textbf{90.0} & \underline{61.1} & 53.2 & \underline{66.1}
        & \underline{74.1} & \underline{56.0} & \underline{44.9} & \underline{56.3}
        & \underline{89.8} & 79.2 & 72.6 & 79.8 \\
        \rowcolor{lightgray}
        \textbf{+ TARS (Ours)}
        & \underline{88.9} & \textbf{62.0} & \textbf{60.3}$_{\textcolor{tarsred}{\uparrow 5.5}}$ & \textbf{68.8}$_{\textcolor{tarsred}{\uparrow 3.1}}$
        & \textbf{75.6} & \textbf{56.5} & \textbf{51.7}$_{\textcolor{tarsred}{\uparrow 8.2}}$ & \textbf{59.3}$_{\textcolor{tarsred}{\uparrow 5.0}}$
        & \textbf{90.1} & \underline{80.1} & \textbf{76.0}$_{\textcolor{tarsred}{\uparrow 2.1}}$ & \textbf{81.3}$_{\textcolor{tarsred}{\uparrow 0.3}}$ \\
        \bottomrule
    \end{tabular}}
    \end{center}
    \vspace{-3.0mm}
\end{table}
\begin{table}[t!]
    \setlength{\tabcolsep}{1.5pt}
    \newcolumntype{M}{>{\centering\arraybackslash}m{3.4em}}
    \caption{
    Domain-specific detection results on IP102. We report precision, recall, and F1 score (\%) at an IoU threshold of $0.5$, averaged over classes within each frequency group and across all classes. SFT baselines and formatting follow Tab.~\ref{tab:open_world_classification}. \textcolor{tarsred}{Red} subscripts show TARS gains in percentage points over the corresponding SFT baseline for Few and All. 
    }
    \vspace{-2.0mm}
    \label{tab:ip102_detection}
    \begin{center}
    \resizebox{\textwidth}{!}{%
    \begin{tabular}{l *{12}{M}}
        \toprule
        \multirow{2}{*}{\textbf{Method}}
        & \multicolumn{4}{c}{\textbf{P@0.5}}
        & \multicolumn{4}{c}{\textbf{R@0.5}}
        & \multicolumn{4}{c}{\textbf{F1@0.5}} \\
        \cmidrule(lr){2-5}
        \cmidrule(lr){6-9}
        \cmidrule(lr){10-13}
        & Many & Med & Few & All
        & Many & Med & Few & All
        & Many & Med & Few & All \\
        \midrule
        Qwen2.5-VL-3B
        & 0.0 & 0.0 & 0.0 & 0.0
        & 0.0 & 0.0 & 0.0 & 0.0
        & 0.0 & 0.0 & 0.0 & 0.0 \\
        \midrule
        CE Loss
        & 66.9 & 40.3 & \underline{23.9} & 49.8
        & 68.9 & 38.5 & \underline{25.6} & 50.0
        & 67.3 & 38.0 & \underline{23.8} & 48.9 \\
        + Vanilla
        & \underline{68.7} & \underline{42.2} & 23.6 & \underline{51.4}
        & \textbf{70.3} & \underline{39.4} & 24.4 & \underline{50.9}
        & \textbf{69.0} & \underline{38.6} & 23.5 & \underline{49.8} \\
        \rowcolor{lightgray}
        \textbf{+ TARS (Ours)}
        & \textbf{68.9} & \textbf{43.3} & \textbf{30.2}$_{\textcolor{tarsred}{\uparrow 6.3}}$ & \textbf{52.7}$_{\textcolor{tarsred}{\uparrow 2.9}}$
        & \underline{70.1} & \textbf{40.4} & \textbf{27.3}$_{\textcolor{tarsred}{\uparrow 1.7}}$ & \textbf{51.6}$_{\textcolor{tarsred}{\uparrow 1.6}}$
        & \underline{68.9} & \textbf{39.5} & \textbf{27.7}$_{\textcolor{tarsred}{\uparrow 3.9}}$ & \textbf{50.7}$_{\textcolor{tarsred}{\uparrow 1.8}}$ \\
        \midrule
        Focal Loss
        & \underline{66.5} & \underline{41.4} & \textbf{39.4} & \underline{51.9}
        & 68.2 & \textbf{41.2} & \underline{27.1} & 51.1
        & 66.6 & \textbf{39.7} & \underline{31.1} & \underline{50.2} \\
        + Vanilla
        & \textbf{68.5} & 41.1 & 27.1 & 51.2
        & \underline{70.5} & 39.9 & 25.0 & \underline{51.2}
        & \underline{69.0} & 38.5 & 25.7 & 50.0 \\
        \rowcolor{lightgray}
        \textbf{+ TARS (Ours)}
        & \textbf{68.5} & \textbf{42.9} & \underline{37.8}$_{\textcolor{tarsred}{\downarrow 1.6}}$ & \textbf{53.2}$_{\textcolor{tarsred}{\uparrow 1.3}}$
        & \textbf{70.6} & \underline{40.2} & \textbf{35.0}$_{\textcolor{tarsred}{\uparrow 7.9}}$ & \textbf{52.6}$_{\textcolor{tarsred}{\uparrow 1.5}}$
        & \textbf{69.1} & \underline{38.9} & \textbf{35.3}$_{\textcolor{tarsred}{\uparrow 4.2}}$ & \textbf{51.4}$_{\textcolor{tarsred}{\uparrow 1.2}}$ \\
        \midrule
        Reweighting
        & 66.7 & \underline{44.0} & 32.7 & 52.4
        & \underline{69.1} & \textbf{43.7} & 27.1 & 52.7
        & 67.2 & \textbf{42.4} & 28.4 & 51.4 \\
        + Vanilla
        & \textbf{68.3} & 43.1 & \underline{37.3} & \underline{53.2}
        & \textbf{69.9} & \underline{42.1} & \underline{36.8} & \textbf{53.4}
        & \textbf{68.6} & \underline{41.1} & \underline{35.9} & \textbf{52.3} \\
        \rowcolor{lightgray}
        \textbf{+ TARS (Ours)}
        & \underline{67.7} & \textbf{44.4} & \textbf{40.8}$_{\textcolor{tarsred}{\uparrow 8.1}}$ & \textbf{53.9}$_{\textcolor{tarsred}{\uparrow 1.5}}$
        & \underline{69.1} & 41.6 & \textbf{38.9}$_{\textcolor{tarsred}{\uparrow 11.8}}$ & \underline{53.0}$_{\textcolor{tarsred}{\uparrow 0.3}}$
        & \underline{67.8} & 41.0 & \textbf{39.0}$_{\textcolor{tarsred}{\uparrow 10.6}}$ & \underline{52.2}$_{\textcolor{tarsred}{\uparrow 0.8}}$ \\
        \midrule
        Sampling
        & \textbf{63.0} & \textbf{44.4} & 39.7 & \underline{51.8}
        & 65.6 & \textbf{42.7} & \underline{39.2} & \underline{52.1}
        & 63.6 & \textbf{42.1} & \underline{37.3} & \underline{50.7} \\
        + Vanilla
        & \underline{62.6} & 43.5 & \underline{41.3} & 51.4
        & \underline{66.0} & 40.6 & 36.2 & 51.0
        & \underline{63.8} & 39.7 & 37.2 & 49.7 \\
        \rowcolor{lightgray}
        \textbf{+ TARS (Ours)}
        & \textbf{63.0} & \underline{44.3} & \textbf{44.9}$_{\textcolor{tarsred}{\uparrow 5.2}}$ & \textbf{52.3}$_{\textcolor{tarsred}{\uparrow 0.5}}$
        & \textbf{66.6} & \underline{42.3} & \textbf{40.8}$_{\textcolor{tarsred}{\uparrow 1.6}}$ & \textbf{52.5}$_{\textcolor{tarsred}{\uparrow 0.4}}$
        & \textbf{64.0} & \underline{41.4} & \textbf{40.8}$_{\textcolor{tarsred}{\uparrow 3.5}}$ & \textbf{51.0}$_{\textcolor{tarsred}{\uparrow 0.3}}$ \\
        \bottomrule
    \end{tabular}}
    \end{center}
    \vspace{-3.0mm}
\end{table}

\paragraph{Domain-specific Classification.}
Tab.~\ref{tab:open_world_classification} reports classification accuracy on the two iNaturalist subsets and CUB-LT, split by class frequency in the training set. The overall gains from vanilla GRPO do not consistently extend to tail classes, whose accuracy often drops below the SFT baseline. In contrast, TARS outperforms vanilla GRPO on both Few and All across all four SFT initializations on all three datasets. Starting from CE loss SFT, TARS improves Few accuracy over vanilla GRPO by 6.3, 3.4, and 2.8 percentage points on iNat-Sciuridae, iNat-Coenagrionidae, and CUB-LT, respectively, while also improving overall accuracy. Across all initializations, these gains come with at most a 1.1-point decrease in Many accuracy relative to vanilla GRPO. TARS also improves Few and All accuracy over every imbalance-aware SFT baseline. For example, on iNat-Coenagrionidae, applying TARS after class-balanced sampling raises Few accuracy from 43.5 to 51.7 and overall accuracy from 54.3 to 59.3. These results show that addressing imbalance during SFT does not eliminate the benefit of tailness-aware rescaling during RL.

\paragraph{Domain-specific Detection.}
Tab.~\ref{tab:ip102_detection} evaluates TARS on the detection benchmark IP102, complementing the classification results. Vanilla GRPO does not consistently improve tail-class performance. With SFT using CE and focal loss, it lowers Few F1 from 23.8 to 23.5 and from 31.1 to 25.7, respectively, while changing overall F1 by less than one percentage point. TARS instead raises Few F1 to 27.7 and 35.3, exceeding vanilla GRPO by 4.2 and 9.6 percentage points. Across all four SFT initializations, TARS outperforms vanilla GRPO on Few F1 and improves All F1 in three cases, with a 0.1-point decrease under class-frequency reweighting. TARS also improves Few and All F1 over all four SFT baselines. Together with the classification results, these findings demonstrate TARS's effectiveness across visual recognition tasks and its complementarity with imbalance-aware SFT. 
\looseness=-1

\begin{table}[t]
\caption{
Comparison with imbalance-aware GRPO variants using the same CE loss SFT checkpoint per dataset. We report accuracy (\%) for classification and F1 (\%) at IoU $0.5$ for detection. Bold and underlined values mark the best and second-best results per column.
}
\vspace{-2.0mm}
\label{tab:grpo_variants_full}
\begin{center}
\footnotesize
\setlength{\tabcolsep}{3.2pt}
\resizebox{\textwidth}{!}{%
\begin{tabular}{l cccc cccc cccc cccc}
\toprule
\multirow{2}{*}{Method} & \multicolumn{4}{c}{iNat-Sciuridae} & \multicolumn{4}{c}{iNat-Coenagrionidae} & \multicolumn{4}{c}{CUB-LT} & \multicolumn{4}{c}{IP102} \\
\cmidrule(lr){2-5} \cmidrule(lr){6-9} \cmidrule(lr){10-13} \cmidrule(lr){14-17}
& Many & Med & Few & All & Many & Med & Few & All & Many & Med & Few & All & Many & Med & Few & All \\
\midrule
    CE Loss 
    & 86.7 & 50.0 & 42.9 & 57.4 
    & 77.8 & 56.9 & 43.5 & 57.0 
    & \underline{92.7} & 77.9 & 65.8 
    & 78.2 & 67.3 & 38.0 & 23.8 & 48.9 \\
    + Vanilla 
    & \underline{88.9} & 54.6 & 42.1 & 59.3 
    & \underline{80.7} & 55.6 & 43.0 & 57.0 
    & 91.6 & \underline{79.6} & 67.9 & 79.2 
    & \textbf{69.0} & \underline{38.6} & 23.5 & \underline{49.8} \\
    + Sampling 
    & 86.7 & 52.8 & \textbf{52.4} & \underline{62.0} 
    & 77.8 & 54.2 & \textbf{47.8} & 57.5 
    & 89.7 & 78.3 & \textbf{71.3} & 79.3 
    & 67.2 & 37.7 & \textbf{31.2} & 49.6 \\
    + DiVE-k 
    & \textbf{90.0} & 50.0 & 45.2 & 59.3 
    & 77.8 & 58.3 & 40.6 & 56.5 
    & \textbf{92.9} & 77.7 & 66.0 & 78.2 
    & 68.4 & 37.7 & 26.6 & 49.5 \\  
    + DISCO 
    & 87.8 & \textbf{57.4} & 43.7 & 60.5 
    & \textbf{82.2} & 55.6 & 44.9 & 58.1 
    & 91.5 & 79.0 & 68.5 & 79.0 
    & 65.8 & 37.1 & 26.3 & 48.2 \\
    + DRPO 
    & 88.4 & \underline{57.0} & 45.3 & 61.2 
    & 80.0 & \textbf{57.7} & 45.0 & \underline{58.3} 
    & 91.5 & \textbf{79.8} & 69.2 & \underline{79.7} 
    & 67.9 & 37.1 & 25.5 & 48.9 \\
    \rowcolor{lightgray}
    \textbf{+ TARS (Ours)} 
    & \underline{88.9} & \textbf{57.4} & \underline{48.4} & \textbf{62.7} 
    & \underline{80.7} & \underline{57.4} & \underline{46.4} & \textbf{59.0} 
    & 91.5 & 79.5 & \underline{70.7} & \textbf{79.9} 
    & \underline{68.9} & \textbf{39.5} & \underline{27.7} & \textbf{50.7} \\
\bottomrule
\end{tabular}}
\end{center}
\vspace{-3.0mm}
\end{table}

\paragraph{Comparison with Imbalance-aware GRPO.}
We compare TARS with class-balanced sampling during GRPO, DiVE-k \citep{kumar2026dive}, DISCO \citep{zhou2025disco}, and DRPO \citep{dai2026qoq}. Class-balanced sampling increases rare-class exposure, while DiVE-k prioritizes difficult queries through hard-negative mining. DISCO and DRPO address domain imbalance during RL, and we adapt both to class imbalance by treating each class as a domain. All methods start from the same CE loss SFT checkpoint. As shown in Tab.~\ref{tab:grpo_variants_full}, class-balanced sampling outperforms TARS on Few in some benchmarks but yields lower All performance across all four. This trade-off may arise because increasing rare-class exposure reduces the relative exposure of frequent classes. DiVE-k does not consistently improve Few performance over vanilla GRPO, suggesting that prioritizing difficult queries alone does not necessarily address the limited exposure of rare classes. DISCO and DRPO improve Few performance over vanilla GRPO on all four benchmarks, but TARS further outperforms both on Few and All. These differences may reflect the granularity of the signals used to allocate RL updates. DISCO assesses prompt-level difficulty, while DRPO uses reward statistics within each batch. In contrast, TARS combines class rarity with the SFT model's recognition error, providing a stable class-level signal across the training data, not each batch. By selectively amplifying learning signals, TARS improves tail-class recognition while maintaining strong overall performance. Additional experiments on a long-tailed PlantDoc variant \citep{singh2020plantdoc} further validate TARS against GRPO baselines on a separate detection dataset (Appendix~\ref{app:plantdoc}).

\begin{figure}[t!]
    \begin{center}
    \includegraphics[width=\linewidth]{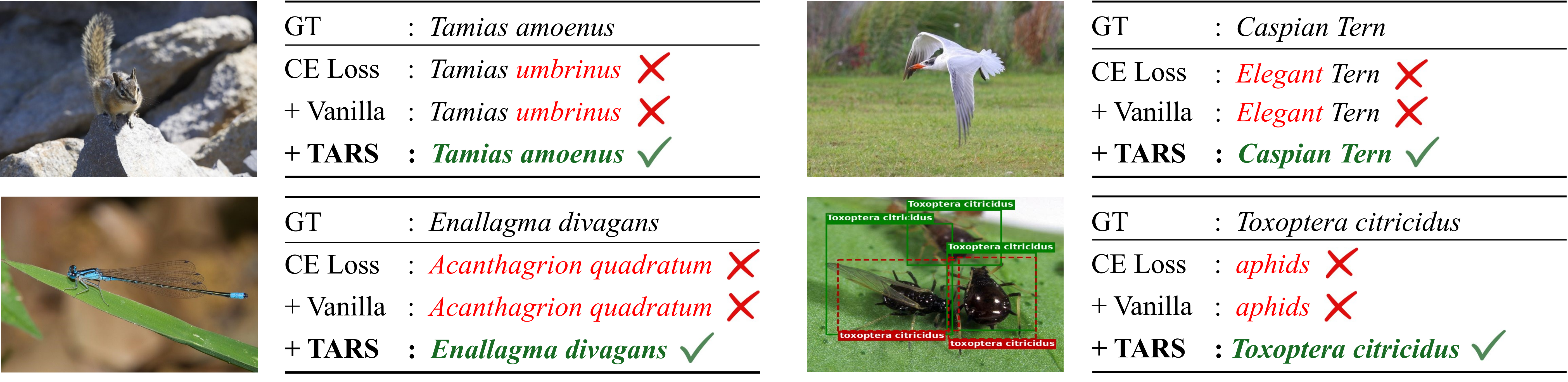}
    \end{center}
    \vspace{-3.0mm}
    \caption{
    Qualitative results on Few classes across all four benchmarks. CE loss SFT and vanilla GRPO predict frequent classes, while TARS predicts the correct tail classes. For IP102, \textcolor{darkgreen}{green} boxes show ground truth and \textcolor{tarsred}{red} boxes show predictions.
    }
    \label{fig:qualitative_cls}
\end{figure}

\paragraph{Qualitative Results.}
Fig.~\ref{fig:qualitative_cls} shows examples from Few classes across all four benchmarks. In the classification examples, the baselines misidentify tail classes as classes that occur frequently in the training data, while TARS predicts the correct labels. In the detection example, the baselines localize the objects but assign them to the frequent class \emph{aphids}, whereas TARS correctly identifies the tail class \emph{Toxoptera citricidus}. These cases illustrate the improvements in tail-class recognition observed in the quantitative results.

\begin{figure}[t!]
    \begin{center}
    \includegraphics[width=\linewidth]{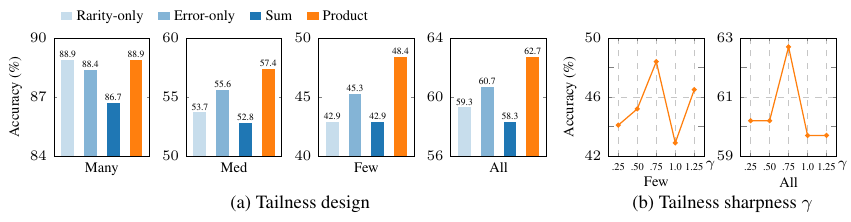}
    \end{center}
    \vspace{-5.0mm}
    \caption{
    Ablations of tailness design and sharpness. (a) Accuracy with tailness based on rarity-only, error-only, their sum, or their product. (b) Few and All accuracy as the exponent $\gamma$ varies.
    }
    \label{fig:ablation}
    \vspace{-3.0mm}
\end{figure}

\subsection{Ablation Studies}
\label{sec:ablation}
We conduct controlled ablations on iNat-Sciuridae, the most imbalanced classification benchmark, to evaluate the key design choices behind TARS. All variants start from the same SFT checkpoint and use the same RL configuration, changing only the component under study.

\paragraph{Tailness Design.}
Fig.~\ref{fig:ablation}(a) compares the product of per-class rarity and recognition error against each signal alone and their sum, with the rest of TARS fixed. The product achieves the best Medium, Few, and All accuracy, reaching 48.4 on Few and 62.7 on All while matching rarity alone on Many. These results support using the product to account for limited exposure while tempering the emphasis on rare classes that are already well recognized.

\paragraph{Tailness Sharpness.}
The exponent $\gamma$ controls the contrast among class weights, with larger values amplifying high-tailness classes more strongly. Fig.~\ref{fig:ablation}(b) shows that stronger amplification does not consistently improve performance. Among the tested values, $\gamma=0.75$ achieves the highest Few and All accuracy, and we use it for classification.

\begin{table}[t!]
    \caption{Comparison between static and dynamic tailness on iNat-Sciuridae. We report accuracy (\%) and additional training time relative to static TARS.}
    \label{tab:static_dynamic_tailness}
    \begin{center}
    \begin{tabular}{lccccc}
        \toprule
        Tailness & Many & Medium & Few & All & Additional time \\
        \midrule
        Dynamic
        & \textbf{90.0} & \textbf{57.4} & 46.8 & 62.3 & +0.98\% \\
        \rowcolor{lightgray}
        \textbf{Static (Ours)}
        & 88.9 & \textbf{57.4} & \textbf{48.4} & \textbf{62.7} & -- \\
        \bottomrule
    \end{tabular}
    \end{center}
    \vspace{-3.0mm}
\end{table}

\paragraph{Static vs. Dynamic Tailness.}
We compare our static tailness with an epoch-wise dynamic variant that recomputes per-class recognition errors on the training set using the current policy after each RL epoch. As shown in Tab.~\ref{tab:static_dynamic_tailness}, dynamic tailness improves Many accuracy but yields lower Few and All accuracy than static tailness. One possible explanation is that dynamic re-estimation reduces the emphasis on rare classes as their training accuracy improves, even when their generalization to unseen examples remains insufficient. In contrast, static tailness maintains consistent emphasis on initially under-recognized rare classes throughout RL. Dynamic re-estimation incurs only a 0.98\% training-time overhead, so computational cost is not the primary reason for favoring static tailness. We therefore adopt static tailness primarily for its better Few and All performance.

Additional ablations in Appendix~\ref{app:add_abl} further validate the one-sided rescaling design and demonstrate TARS's effectiveness on a newer and larger LVLM, Qwen3-VL-8B \citep{bai2025qwen3}.

\section{Limitations and Discussion}
TARS inherits GRPO's limitation that identical rollout rewards yield zero advantages, leaving no learning signals to amplify. However, our RL training starts from an SFT checkpoint with nontrivial rare-class recognition capability. On iNat-Sciuridae, 59\% of Few-class yield both correct and incorrect rollouts under the initial SFT policy, providing informative advantages for TARS to amplify. Another limitation concerns the reliability of per-class recognition errors estimated from the initial SFT model, particularly for extremely rare classes. While recognition error is a component of tailness, it provides useful information for refining the learning emphasis. Fig.~\ref{fig:ablation}(a) demonstrates its benefit over rarity-only weighting, although robustness to estimation noise remains to be investigated. Moreover, domain-specific fine-tuning may also degrade an LVLM's general-purpose capabilities. However, our evaluation on MMBench (Appendix~\ref{general_purpose}) shows only a modest performance decrease after fine-tuning. Finally, TARS currently relies on explicit class labels, but its underlying principle of compensating for imbalanced group exposure could potentially extend to domains or tasks with appropriate definitions of group membership, exposure, and initial model performance.
\looseness=-1

\section{Conclusion}
We studied domain-specific visual recognition with LVLMs under limited and imbalanced training data and showed that vanilla GRPO's gains do not consistently extend to tail classes. Its within-query advantage normalization does not explicitly account for class frequency or class-level recognition performance. We proposed TARS, which rescales GRPO advantages using tailness, a per-class measure combining class rarity with the SFT model's recognition error. Tailness is estimated once before RL and held fixed, requiring no additional inference for tailness estimation during training. Across four classification and detection benchmarks, TARS consistently improves tail-class performance over SFT and vanilla GRPO, while achieving competitive results on tail classes and the highest overall performance among the compared GRPO methods. It also provides further gains on top of imbalance-aware SFT methods, demonstrating their complementarity.

\newpage

\subsection*{AI use statement}

In this work, we used generative AI tools to aid and polish the writing of the manuscript. Additionally, we used generative AI tools for retrieval and discovery, such as identifying potentially relevant related work. We have not used generative AI tools for other tasks with required disclosure, and the rest of the required disclosure tasks are not applicable to this work. We have reviewed all AI-assisted work. In particular, all AI-assisted writing was reviewed and edited by the authors, and literature identified with AI assistance was independently verified against the original sources before being cited. We take responsibility for the final content of this work, including text, claims or artifacts produced with the aid of generative AI.

\bibliography{iclr2027_conference}
\bibliographystyle{iclr2027_conference}

\newpage
\appendix

\section{Additional Implementation Details}
\label{app:additional}

\subsection{Dataset statistics and details}
\label{app:dataset_details}

\begin{table}[h]
\caption{Statistics of the four domain-specific benchmarks. IF denotes the imbalance factor, computed as the ratio of the largest to the smallest class frequency in the training set.}
\label{tab:dataset_stats}
\begin{center}
\begin{tabular}{llrrr}
\toprule
Dataset & Task & Classes & Train Images & IF \\
\midrule
iNat-Sciuridae & Classification & 36 & 4,996 & 90.9 \\
iNat-Coenagrionidae & Classification & 62 & 4,042 & 33.9 \\
CUB-LT & Classification & 200 & 2,945 & 14.3 \\
IP102 & Detection & 97 & 15,178 & 1,174 \\
\bottomrule
\end{tabular}
\end{center}
\end{table}

\subsection{Training settings}
We use AdamW \citep{loshchilov2019decoupled} with a cosine learning-rate schedule and a warmup ratio of $0.03$ for both SFT and RL. Tab.~\ref{tab:training_settings} lists the benchmark-specific settings, which are shared across methods within each benchmark. Images use a pixel budget ranging from $512\times28\times28$ to $1280\times28\times28$ during SFT, RL, and evaluation. SFT uses a global batch size of $64$ examples, while each RL step uses $8$ queries with $8$ rollouts each. We train with DeepSpeed ZeRO-2 \citep{rajbhandari2020zero}. \looseness=-1

\begin{table}[H]
    \caption{Training settings for each benchmark. $\gamma_{\mathrm{f}}$ and $k$ are the focal loss and class-frequency reweighting exponents, respectively.}
    \label{tab:training_settings}
    \begin{center}
    \begin{tabular}{lcccccc}
        \toprule
        Benchmark & SFT LR & SFT epochs & RL LR & RL epochs & $\gamma_{\mathrm{f}}$ & $k$ \\
        \midrule
        iNat-Sciuridae
        & $5\times10^{-6}$ & 5 & $5\times10^{-6}$ & 3 & 1.0 & 0.25 \\
        iNat-Coenagrionidae
        & $5\times10^{-6}$ & 5 & $5\times10^{-6}$ & 3 & 2.0 & 0.25 \\
        CUB-LT
        & $5\times10^{-6}$ & 5 & $1\times10^{-6}$ & 3 & 1.0 & 0.25 \\
        IP102
        & $5\times10^{-5}$ & 3 & $5\times10^{-6}$ & 1 & 1.0 & 0.25 \\
        \bottomrule
    \end{tabular}
    \end{center}
\end{table}

\newpage
\subsection{Imbalance-aware SFT Formulation for LVLMs}
Cross-entropy, focal loss \citep{lin2017focal}, and class-frequency reweighting \citep{du2023global} differ in their loss weights, while class-balanced sampling \citep{kang2020decoupling} retains cross-entropy and changes the sampling distribution. The loss-based SFT baselines differ in how they weight answer tokens. Let $\mathcal{B}$ denote an SFT mini-batch, $x_i$ the image and question of sample $i$, and $y_i$ its ground-truth answer. We compute the loss over the $T_i$ tokens in $y_i$. Let $p_{i,t}=\pi_{\theta}(y_{i,t}\mid x_i,y_{i,<t})$ denote the probability assigned to the $t$-th answer token given the input and preceding ground-truth tokens. The loss is
\begin{equation}
\mathcal{L} =
-\frac{1}{\sum_{i \in \mathcal{B}} T_i}
\sum_{i \in \mathcal{B}}\sum_{t=1}^{T_i}
W_{i,t}\log p_{i,t}.
\label{eq:sft_general}
\end{equation}
For the loss-based classification baselines, we instantiate the token weights in Eq.~\ref{eq:sft_general} as follows:
\begin{equation}
W_{i,t}=
\begin{cases}
1, & \text{cross-entropy},\\[2pt]
(1-p_{i,t})^{\gamma_{\mathrm{f}}}, & \text{focal loss},\\[2pt]
w'_{c_i}, & \text{class-frequency reweighting},
\end{cases}
\qquad
w'_c =
\frac{|\mathcal{C}|\,n_c^{-k}}
{\sum_{j \in \mathcal{C}} n_j^{-k}},
\label{eq:sft_weights}
\end{equation}
where $n_c$ is the training image count defined in Eq.~\ref{eq:tailness}. Focal loss reduces the contribution of tokens that the model predicts with high confidence, while class-frequency reweighting gives greater weight to under-represented classes. The values of $\gamma_{\mathrm{f}}$ and $k$ are listed in Tab.~\ref{tab:training_settings}.

Class-balanced sampling is a separate data-level baseline. It selects a class uniformly and then samples an image uniformly from that class for classification. Sampling uses replacement and retains the original number of training examples per epoch, keeping the number of optimization steps unchanged.
\looseness=-1

For detection, focal loss is applied to all answer tokens, including class labels, coordinates, and JSON structure. Class-frequency reweighting applies $w'_c$ to tokens corresponding to the class label and bounding box of each object. Class-balanced sampling assigns each image a weight based on the inverse frequency of the rarest class it contains and samples images accordingly.

\newpage
\section{Additional Ablation Studies}
\label{app:add_abl}

\subsection{One-sided Rescaling}
\begin{figure}[h]
\begin{center}
\includegraphics[width=0.55\columnwidth]{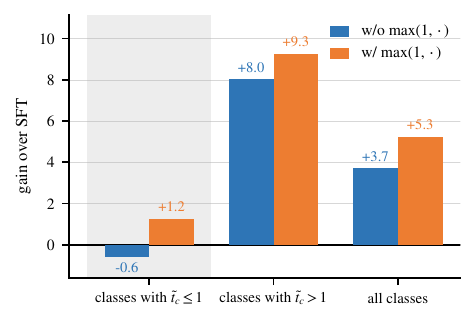}
\vspace{-4.0mm}
\end{center}
\caption{Effect of one-sided rescaling on iNat-Sciuridae. Accuracy gains over the CE loss SFT checkpoint in percentage points, grouped by calibrated tailness.}
\label{fig:one_sided_rescaling}
\end{figure}

We ablate the one-sided design of TARS by replacing $w_c=\max(1,\tilde{t}_c)$ with $w_c=\tilde{t}_c$, while keeping all other settings unchanged. This reduces the magnitude of advantages for classes with $\tilde{t}_c\le1$ below the standard GRPO scale. Classes with $\tilde{t}_c>1$ receive the same amplification in both variants. As shown in Fig.~\ref{fig:one_sided_rescaling}, removing one-sided rescaling changes the gain over the SFT checkpoint from $+1.2$ to $-0.6$ points for classes with $\tilde{t}_c\leq1$. Performance also decreases for classes with $\tilde{t}_c>1$, resulting in lower overall accuracy. These results support the one-sided design of TARS, which preserves the standard GRPO advantage scale for lower-tailness classes while selectively amplifying higher-tailness classes.

\subsection{Newer and Larger LVLM}
To examine whether the benefits of TARS extend to newer and larger LVLMs, we conduct an additional ablation using Qwen3-VL-8B \citep{bai2025qwen3} on iNat-Sciuridae, following the experimental setting of our ablation studies. We compare TARS with vanilla GRPO, class-balanced sampling, DiVE-k, DISCO, and DRPO, all initialized from the same CE loss SFT checkpoint.

\begin{table}[h]
\caption{Classification accuracy (\%) on iNat-Sciuridae using Qwen3-VL-8B.}
\label{tab:qwen3vl}
\begin{center}
\begin{tabular}{lcccc}
\toprule
Method & Many & Med & Few & All \\
\midrule
Qwen3-VL-8B (zero-shot) & 53.3 & 11.1 & 16.7 & 25.0 \\
CE Loss & 83.3 & 58.3 & 52.4 & 63.0 \\
+ Vanilla & \textbf{90.0} & 66.7 & 45.2 & 64.8 \\
+ Sampling & \underline{86.7} & 63.9 & \textbf{59.5} & \underline{68.5} \\
+ DiVE-k   & 83.3 & 61.1 & 45.2 & 61.1 \\
+ DISCO    & \underline{86.7} & \underline{69.4} & 52.4 & 67.6 \\ 
+ DRPO     & 83.3 & \underline{69.4} & 50.0 & 65.7 \\
\rowcolor{lightgray}
+ TARS (Ours) & \textbf{90.0} & \textbf{72.2} & \underline{57.1} & \textbf{71.3} \\
\bottomrule
\end{tabular}
\end{center}
\end{table}

As shown in Tab.~\ref{tab:qwen3vl}, vanilla GRPO reduces Few accuracy from 52.4 to 45.2 compared with CE loss SFT. DiVE-k shows no improvement in Few accuracy over vanilla GRPO, while DISCO and DRPO achieve lower overall accuracy than TARS. Class-balanced sampling achieves the highest Few accuracy (59.5), but yields lower Many and Medium accuracy than TARS. In contrast, TARS improves Few accuracy by 11.9 points and overall accuracy by 6.5 points over vanilla GRPO, while maintaining Many accuracy. It achieves the highest overall accuracy (71.3) among all compared methods, demonstrating that TARS's benefits extend to a newer and larger LVLM.

\newpage
\section{Further Analysis}

\subsection{Additional Detection Benchmark: PlantDoc}
\label{app:plantdoc}

To further evaluate TARS on domain-specific object detection, we conduct additional experiments on PlantDoc \citep{singh2020plantdoc}, a dataset for plant disease detection. The original PlantDoc training set contains 2,128 images across 30 classes, with an imbalance factor (IF) of 164. However, we merge one duplicated class (`soyabean leaf' and `soybean leaf') and remove two classes absent from the evaluation split (`tomato two spotted spider mites leaf' and `potato leaf'), leaving 2,114 training images across 27 classes with an imbalance factor (IF) of 4.2. Since this does not satisfy our imbalance criterion (IF $> 10$), we construct a long-tailed variant by subsampling the training images, following the approach of constructing CUB-LT from CUB-200-2011. The resulting dataset contains 1,224 training images with an IF of 32.8.

We use Qwen2.5-VL-3B-Instruct and compare TARS with vanilla GRPO, DiVE-k, and class-balanced sampling during GRPO, all initialized from the same CE loss SFT checkpoint. For PlantDoc, we adapt the detection prompt in Tab.~\ref{tab:query-examples} to refer to plant diseases instead of pests. SFT is trained for 3 epochs with a learning rate of $5 \times 10^{-5}$, while RL uses 3 epochs with a learning rate of $5 \times 10^{-6}$. We follow the same detection evaluation protocol and class-frequency threshold used for IP102 and report F1 scores at an IoU threshold of 0.5 on the PlantDoc validation split.

\begin{table}[h]
\caption{Detection performance (F1@0.5, \%) on the long-tailed PlantDoc variant. Many, Medium, and Few denote class-frequency groups, and All denotes the overall class-averaged F1 score.}
\label{tab:plantdoc}
\begin{center}
\begin{tabular}{lcccc}
\toprule
Method & Many & Med & Few & All \\
\midrule
CE Loss & 36.4 & 22.2 & 18.2 & 22.7 \\
+ Vanilla & 38.4 & \textbf{27.1} & 20.4 & 26.1 \\
+ DiVE-k & 37.9 & 24.0 & 20.1 & 24.5 \\
+ Sampling & \textbf{47.9} & 25.5 & \underline{22.0} & \underline{27.4} \\
\rowcolor{lightgray}
+ TARS (Ours) & \underline{42.4} & \underline{26.5} & \textbf{24.0} & \textbf{27.8} \\
\bottomrule
\end{tabular}
\end{center}
\end{table}

As shown in Tab.~\ref{tab:plantdoc}, DiVE-k yields lower F1 scores than vanilla GRPO across all frequency groups, suggesting that prioritizing difficult queries alone does not necessarily address class imbalance. Class-balanced sampling achieves the highest Many F1 and improves Few F1 over vanilla GRPO, but its gains in Few and overall performance remain smaller than those of TARS. In contrast, TARS achieves the highest Few and overall F1 scores, outperforming vanilla GRPO by 3.6 and 1.7 percentage points, respectively. TARS also improves performance across all frequency groups relative to CE loss SFT. These results further demonstrate the effectiveness of tailness-aware advantage rescaling in domain-specific detection under limited and imbalanced training data.

\newpage
\subsection{General-Purpose Capability Retention}
\label{general_purpose}
Domain-specific fine-tuning may degrade an LVLM's general-purpose capabilities. To examine this potential trade-off, we evaluate the original Qwen2.5-VL-3B-Instruct model, the CE loss SFT checkpoints, and the vanilla GRPO and TARS models initialized from those checkpoints on MMBench-EN \citep{liu2024mmbench} (test v1.1), which assesses general-purpose multimodal perception and reasoning capabilities.

\begin{table}[h]
    \caption{General-purpose multimodal performance on MMBench-EN (test v1.1) after domain-specific fine-tuning.}
    \label{tab:mmbench}
    \begin{center}
    \begin{tabular}{llc}
    \toprule
    Training Dataset & Method & MMBench (\%) \\
    \midrule
    -- & Qwen2.5-VL-3B & 77.5 \\
    \midrule
    \multirow{3}{*}{iNat-Sciuridae}
        & CE Loss       & 76.2 \\
        & + Vanilla     & 76.2 \\
        & \cellcolor{lightgray}+ TARS (Ours) & \cellcolor{lightgray}76.2 \\
    \midrule
    \multirow{3}{*}{iNat-Coenagrionidae}
        & CE Loss       & 76.5 \\
        & + Vanilla     & 76.3 \\
        & \cellcolor{lightgray}+ TARS (Ours) & \cellcolor{lightgray}76.4 \\
    \midrule
    \multirow{3}{*}{CUB-LT}
        & CE Loss       & 75.5 \\
        & + Vanilla     & 75.5 \\
        & \cellcolor{lightgray}+ TARS (Ours) & \cellcolor{lightgray}75.5 \\
    \midrule
    \multirow{3}{*}{IP102}
        & CE Loss       & 76.3 \\
        & + Vanilla     & 76.8 \\
        & \cellcolor{lightgray}+ TARS (Ours) & \cellcolor{lightgray}76.8 \\
    \bottomrule
    \end{tabular}
    \end{center}
\end{table}

As shown in Tab.~\ref{tab:mmbench}, domain-specific fine-tuning leads to a modest performance decrease of at most 2.0 percentage points compared with the original model. Importantly, TARS matches or slightly exceeds vanilla GRPO on MMBench across all four datasets. These results suggest that the models largely retain their general-purpose multimodal capabilities after fine-tuning, while TARS improves domain-specific recognition without introducing additional degradation on MMBench.

\newpage
\section{Tailness Statistics}
\label{app:tailness_stats}
\begin{figure}[h]
\begin{center}
\includegraphics[width=\columnwidth]{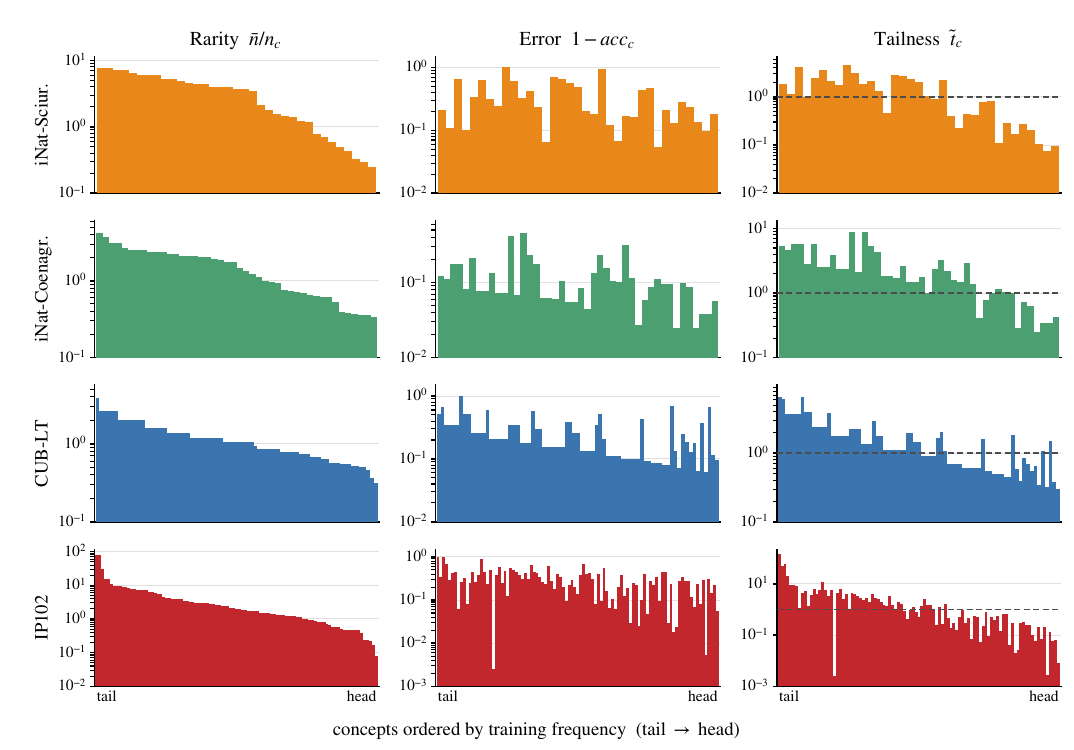}
\end{center}
\caption{Rarity, error, and the calibrated tailness on the training set of each benchmark. Classes are sorted from the rarest to the most frequent, and all $y$-axes are in log scale. TARS amplifies advantages for classes above the dashed line at $\tilde{t}_c=1$ and retains the vanilla GRPO advantage scale for the remaining classes.
\looseness=-1}
\label{fig:tailness_components}
\end{figure}

Fig.~\ref{fig:tailness_components} visualizes the two factors of tailness and the resulting calibrated tailness on all four benchmarks. Given their highly skewed distributions, we plot these quantities on logarithmic scales. Rarity is inversely proportional to class frequency, so a weight built on it alone would reduce to frequency-based reweighting. However, the SFT model’s per-class recognition error does not show a strictly monotonic trend with class frequency. Thus, the product is used to capture both class rarity and recognition error, giving greater emphasis to classes that are under-represented and poorly recognized. Tailness therefore follows neither factor on its own, and the classes it lifts are the ones actually starved of signal, consistent with the tailness design ablation.
\looseness=-1

\end{document}